\documentclass[sigconf]{acmart}
\usepackage[utf8]{inputenc}
\usepackage{float}
\usepackage{booktabs}
\setcopyright{none}
\renewcommand\footnotetextcopyrightpermission[1]{}
\title{Code Completion using Neural Attention and Byte Pair Encoding}

 \author{Youri Arkesteijn}
\affiliation{%
 \institution{y.e.s.s.arkesteijn@student.tudelft.nl}
 \city{Delft University of Technology}}
 
\author{Nikhil Saldanha}
\affiliation{%
 \institution{n.l.saldanha@student.tudelft.nl}
 \city{Delft University of Technology}}
 
 \author{Bastijn Kostense}
\affiliation{%
 \institution{b.kostense@student.tudelft.nl}
 \city{Delft University of Technology}}

\begin{document}
\maketitle

\section{Abstract}
\label{sec:abst}
In this paper, we aim to do code completion based on implementing a Neural Network from Li et. al. \cite{li2017code}. Our contribution is that we use an encoding that is in-between character and word encoding called Byte Pair Encoding (BPE) \cite{shibata1999byte}. We use this on the source code files treating them as natural text without first going through the abstract syntax tree (AST). We have implemented two models: an attention-enhanced LSTM and a pointer network, where the pointer network was originally introduced to solve out of vocabulary problems. We are interested to see if BPE can replace the need for the pointer network for code completion.

\section{Introduction}
Code completion is when the IDE suggests the next probable code tokens based on existing code in the context. Research in the field of code completion is important as programmers are becoming vastly more numerous, and code completion tools are one of the most important tools for a programmer to reduce workload. Therefore, the more accurate code completion tools are, the less work a programmer has to put in to produce a certain piece of code. Since the amount of programmers is increasing fast, accurate code completion tools could save tremendous amounts of work.

\subsection{Out of Vocabulary Problem in Source Code}
Source code has a non-fixed dictionary size, and variance in method, class, and variable names. This means that many word based encodings will provide a UNK token. We call this an out of vocabulary (OoV) prediction. In most works, the OoV problem is handled by reducing the size of the vocabulary by using the most frequent \(K\) words from the corpus and replacing OoV words with a \(UNK\) token. This reduces the computational complexity of computing a high-dimensional Softmax, which is generally the last layer of neural language models. However, this introduces another problem of predicting that the next token is \(UNK\), which is not so useful.

One of the major contributions of \cite{li2017code} was the better prediction of OoV words in code completion using the pointer-mixture network. This pointer-mixture network learns to predict tokens in the local context, where the attention-enhanced LSTM would fail to predict a token in vocabulary.

\subsection{Our Contribution}
\label{subsec:ourcontr}
We argue that predicting from local context is not a complete solution to the OoV problem because the predictions are limited by the size of the input context. We also argue that the increase in accuracy and OoV rate that the authors of \cite{li2017code} reported are seen only because the pointer network gets chances to predict unknown tokens, whereas in the attention-enhanced LSTM unknown tokens are treated as wrong predictions. This means that the pointer network is not tackling the OoV problem head-on; rather, it is reducing the occurrence of the unknown token predictions.

\textbf{The main contribution of this work is the following:}\\
To tackle the OoV problem head-on, we propose to use Byte Pair Encoding on the input data which can learn to encode rare and even unknown tokens as sequences of sub-word units \cite{sennrich-etal-2016-neural}. We believe that the compact vocabulary generated by BPE will result in less frequent OoV predictions and hence obviates the need for the pointer-mixture network. The pointer-mixture network trained on BPE encoded tokens could even reduce the accuracy of the predictions if it has a bias towards predicting words in a local context. We will revisit this in the Results section.

\section{Related work}
In this paper, we aim to do code completion by implementing a variation on \cite{li2017code}. This paper was authored by Jian Li, Yue Wang, Irwin King, and Michael R. Lyu. All authors are affiliated with The Chinese University of Hong Kong. The paper was published in 2017 and was cited 14 times on Google Scholar at the time of writing.

\begin{figure}[!htb]
    \centering
    \includegraphics[width=0.5\textwidth]{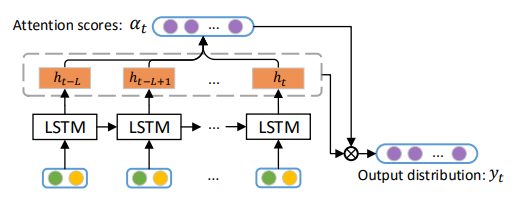}
    \caption{Illustration of the attention-enhanced LSTM taken from \cite{li2017code}.}
    \label{fig:aelstm}
\end{figure}

The attention-enhanced LSTM from \cite{li2017code} is given in figure \ref{fig:aelstm}. The program that is fed into the network is first converted to its abstract syntax tree (AST). Every node in the AST consists of a tuple $(T_i, V_i)$, where $T_i$ is the type of node $i$ and $V_i$ is the value of node $i$. These AST nodes are encoded, embedded, and then fed into the LSTM, where the green dots represent $T_i$ and the yellow dots represent $V_i$. The LSTM then outputs a vector. This is element-wise multiplied by the attention scores $\alpha_t$ to create an output distribution of $y_t$.

\begin{figure}[!htb]
    \centering
    \includegraphics[width=0.5\textwidth]{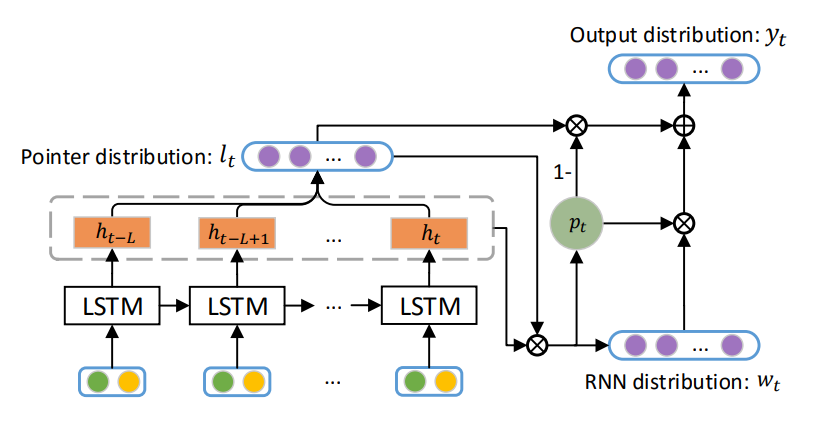}
    \caption{Illustration of the pointer mixture network taken from \cite{li2017code}.}
    \label{fig:pmm}
\end{figure}

However, this is not the final model that is used for code completion in \cite{li2017code}. To balance predicting a token from the global context or the local context, the pointer mixture network is used. The authors state that their pointer mixture network consists of two components: the global RNN component and the local pointer component. They combine and balance these components using a controller. The global component uses the attention-enhanced LSTM from figure 1 to predict a token from a predefined vocabulary, resulting in the RNN distribution $w_t$ from figure 2. This predefined global vocabulary is created by taking the $K$ most frequent tokens. All Out-of-Vocabulary (OoV) tokens are replaced with the special \textit{UNK} token. The local components use a pointer to point to a token in the local context by using the pointer distribution $l_t$ from the global component.

The controller, indicated by $p_t$, is a value between 0 and 1, such that it balances the two distributions $l_t$ and $w_t$ based on either the global or local context. This controller multiplies its value $p_t$ with the global component $w_t$ and $1 - p_t$ with the local component $l_t$. These results are then added up to achieve the final output distribution $y_t$. For validation purposes, we can then compare $y_t$ to the original label of the token sequence to assess the quality of the model.

\section{Data acquisition and generation}
\label{sec:datacqui}
We used SRI lab's 150k Python Dataset\footnote{\url{https://www.sri.inf.ethz.ch/py150}} as a basis for our prediction task. 
Figure \ref{fig:pmm} depicts an overview of our data processing pipeline. We will discuss this process with more detail now. How to run the data acquisition and generation can be found in our README \footnote{\url{https://github.com/serg-ml4se-2019/group6-code-completion}}.

\begin{figure}[!htb]
    \centering
    \includegraphics[width=0.4\textwidth]{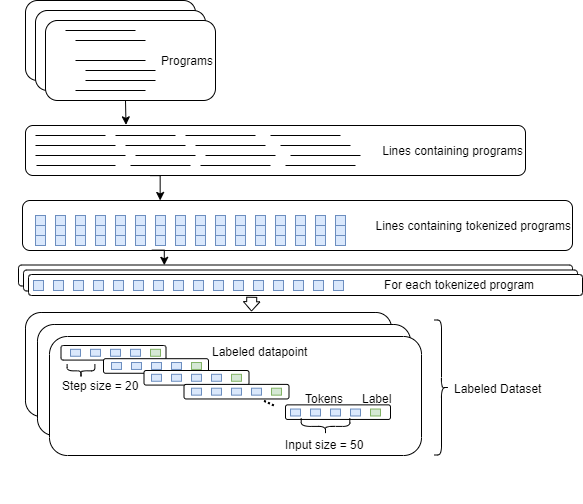}
    \caption{Overview of our data processing pipeline. Here lines represent lines of code. squares are tokens, where blue squares are input and green squares are labels. for more details please read Data acquisition and generation}
    \label{fig:pmm}
\end{figure}

\subsection{Processing}
We take the Python files in the 150k Python Dataset and remove all whitespaces and newlines, such that all expressions are separated by a single space. This results in each Python file of source code being on just a single line. We then create a file containing all these lines, which we use for the encoding task.

\subsection{Encoding}
\label{subs:enc}
Machine learning for code completion often suffers from the Out of Vocabulary Problem because source code contains rare and unseen words which the model never learns to predict. We propose to alleviate this by using BPE \footnote{\url{https://github.com/google/sentencepiece}} to encode our dataset which now contains sub-word tokens instead of word tokens to drastically reduce the size of the vocabulary, making the downstream learning task more tractable. Our choice of using the plain text files instead of the ASTs to encode the data differs from the implementation in \cite{li2017code}. We have chosen this approach since we encode sub-tokens with Byte Pair Encoding, therefore just using the plain text would be sufficient. We also believe that code has patterns similar to natural language since it is an act of communication. This is nicely summarized by the \textbf{Naturalness of Code Hypothesis} as mentioned in \cite{allamanis2018survey}:
\begin{quote}
    "The naturalness hypothesis holds that, because coding is an act of communication, one might expect large code corpora to have rich patterns, similar to natural language"
\end{quote}
Another perspective on the naturalness hypothesis is that both code and language share two important characteristics, namely the presence of syntax in the form of grammar and structure invalid sequences, and semantically that words have a meaning and can refer to each other.

Back to our encoding model, we train a BPE model with a vocabulary size of 8k from our file with 150k files of source code. We then used this model to encode the entire train set as described below.

\subsection{Generation of labelled training data}
\label{subsec:generation}
To generate the labelled training data, we iterate over the source code with a sliding window of $N + 1$ tokens (with a step size of $S$). These are encoded using the trained model from above. The first $N$ encoded tokens are used as the training input parameters and the last encoded token is used as the label. We also include a script to replicate this in our repository.

The final training data that we used to train the model was done by using a sliding window of 50 with a step size of 20, which we use as input to the embedding layer of our model. This is shown in figure \ref{fig:pmm}.

Unfortunately, due to space constraints, we were unable to generate the entire training data. for more on this please see  section \ref{sec:refl}.

\section{Models}
\label{sec:mod}
In this section, we will discuss our two implementations.

\subsection{Attention-enhanced LSTM}
\label{sec:modattn}
\begin{figure}[!htb]
    \centering
    \includegraphics[width=0.5\textwidth]{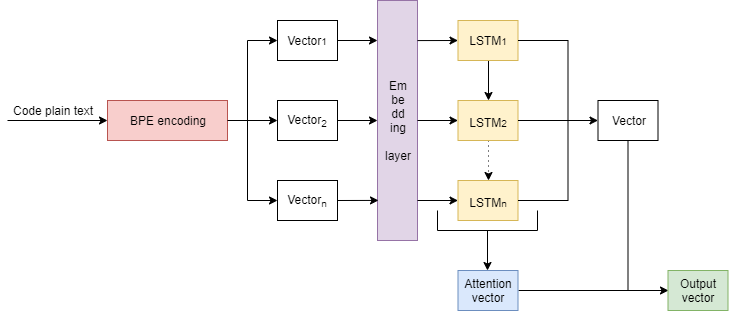}
    \caption{Illustration of our implementation of the attention-enhanced LSTM.}
    \label{fig:ouraelstm}
\end{figure}

The first model we implemented is an attention-enhanced LSTM. A schematic overview of our model can be found in figure \ref{fig:ouraelstm}.

After encoding the plain text as described in subsection \ref{subs:enc}, the input is embedded into a vector of length 256. The weights of the embedding are initialized randomly and learnt jointly with the rest of the model. The embedded vector is fed into an attention-enhanced LSTM with a batch size of 128, a learning rate of 0.001, Adam optimizer, and a gradient clipping of 5. We train this LSTM for 8 epochs. The LSTM consists of a single hidden layer of size 128 and outputs a vector. This vector is then multiplied with the attention vector, resulting in the final output vector. Li et. al. \cite{li2017code} also included a learning rate decay of 0.6, but given the usage of Adam, this is likely to make the results worse so we chose not to include this in our model.

\subsection{Pointer network}
\begin{figure}[!htb]
    \centering
    \includegraphics[width=0.5\textwidth]{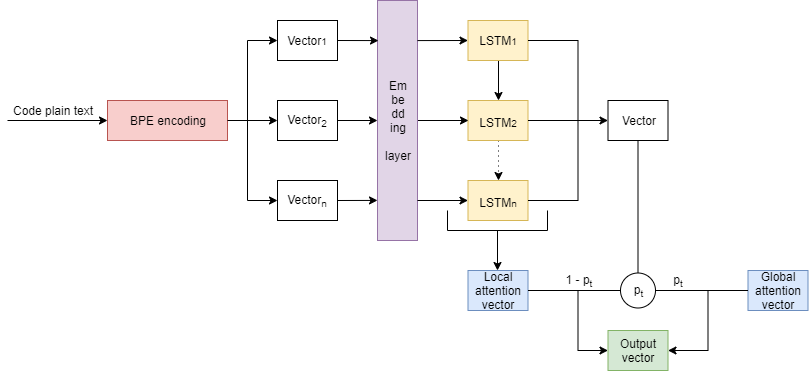}
    \caption{Illustration of our implementation of the pointer network.}
    \label{fig:ourpn}
\end{figure}

In this subsection, we will discuss the second model we implemented; a pointer network. An overview of our model can be found in figure \ref{fig:ourpn}.

As you can see, the difference between this model and the previous model we implemented is that we added a pointer component with a controller that balances predictions from the local and global scope. It was unclear in the original paper about how they computed the loss when the global component dominates the vocabulary component. As we understood it, we construct the label by finding the token the local context vector points to and then create a one-hot encoding of vocabulary vector with a 1 in the location corresponding to the token and the simply evaluate this as we would with an in vocabulary prediction.

However, as stated in section \ref{subsec:ourcontr}, it is important to keep in mind that this pointer network might reduce the accuracy from our model since the goal of the pointer network is to fix the OoV problem. However, the Byte Pair Encoding should also take care of this problem.

Note that the hyperparameters for this network are the same as the Attention-Enhanced LSTM describe in section \ref{sec:modattn}.

\section{Evaluation}
In this section, we will evaluate our implementations by presenting and comparing the results. Afterwards, we will discuss certain points of our implementation that might have affected these results.

\subsection{Results}
The results from our models are shown in table \ref{tab:results}.

\begin{table}[H]
\begin{tabular}{@{}l|l@{}}
\toprule
 & Accuracy \\ \midrule
Attention Enhanced LSTM with BPE (ours) & 69.94\% \\ \midrule
Pointer Mixture Network with BPE (ours) & 58.04\% \\ \midrule
Attention Enhanced LSTM(Li et. al. 2017) & 80.6\% \\ \midrule
Pointer Mixture Network(Li et. al. 2017) & 81.0\% \\ \bottomrule
\end{tabular}
\caption{Comparison of metrics between our implementation of Attention Enhanced LSTM and the Pointer Mixture Network and that of Li et. al. 2017 \cite{li2017code}}
\label{tab:results}
\end{table}

\subsection{Discussion \& Conclusion}
As shown in table \ref{tab:results}, our attention enhanced LSTM vastly outperforms our pointer mixture network in terms of accuracy. This confirms our hypothesis that we stated in subsection \ref{subsec:ourcontr} where we state that the pointer component could reduce the accuracy and that the BPE encoding makes the pointer-mixture network obsolete. This result also hints towards BPE being able to tackle the OoV problem better than a pointer-mixture network.

In the original paper by Li et al. 2017 \cite{li2017code}, the authors train both the attention enhanced LSTM and the Pointer Mixture Network on encoded ASTs of 150k Python Dataset, whereas in our implementation we use BPE on the plain text of the data. Both models from Li et al. 2017 \cite{li2017code} outperform our attention enhanced LSTM. However, this is not due to an inferior implementation. We discuss some of the reasons why our accuracy is worse than the models from the original paper in section \ref{sec:refl}. If we can tackle these problems, we suspect that our attention enhanced LSTM with Byte Pair Encoding could outperform the models from Li et al. 2017 \cite{li2017code}.

Despite having some small problems with the generation of our training data, the model performs surprisingly well. To reflect on the question we asked ourselves in section \ref{sec:abst} where we were wondering whether BPE could replace the need for pointer networks in code completion, we believe that BPE is indeed a suitable replacement.

\section{Reflections}
\label{sec:refl}
There were several things that we could have done that would have likely improve the results we have gotten.
\begin{enumerate}
    \item The sample generation for training and evaluation never fully completed and thus we only trained on a less than 5\% of all data. This happened because we did not expect the amount of data to explode as it did; the training set is almost 500 MB, but the fully processed result would be at least around 40 GB. This happens because of the way we generate our labelled data as described in subsection \ref{subsec:generation}, resulting in a large increase of data. Also, training on all this data could have taken weeks, meaning that this amount of data would have been infeasible in the scope of this project. However, training on all this data could significantly increase the accuracy of our models.
    \item We only ran our training for 8 epochs, because otherwise, the training process would have been infeasible. One of the reasons why we believe that the training is so slow is because we could not train the model on multiple GPUs since our model is not optimized for multi-GPU processing. If we were to implement the GPU optimized code, it will allow us to train for more epochs in a reasonable amount of time which will likely have a positive impact on the accuracy.
    \item We suspect that our implementation of the pointer network is not the most ideal since the paper was not very clear about how to compare the labelled data with the prediction from the model.
    \item The models were trained on Amazon AWS SageMaker on an ml.p2.xlarge instance with 4 vCPUs, 1 Tesla K80 GPU with 12GB of memory. It took about 24 hours to train each model. In total, we spent about \$102, which we believe is quite reasonable.
    \item The data might have been skewed because of the way we create our labelled data as described in subsection \ref{subsec:generation}. Since we often have a case where there are not 50 tokens available at the end of the document, we often have to append 0s to the vector. However, the case where we have to prepend 0s because a prediction occurs at the very start of a document never occurs. This could have a negative impact when using our model on tokens at the beginning of a file.
\end{enumerate}

% 5. Positive: we worked closely/tightly together, which was pleasant.
% 6. How quick it runs on your system

\section{Future work}
Since we did not train on the complete data set nor for a larger number of epochs, future work might be to train the model for a larger amount of epochs on the complete data set and see how the model performs. With this work, we might be able to draw even more reliable conclusions on the suitability for BPE to tackle the Out of Vocabulary prediction in code completion.

Possible further future research also includes a look into better sample generation, because currently there are no samples generated with nothing in the beginning and then only a small amount of meaning full tokens at the end, resulting in the fact that application of this model in the real world will likely fail unless there is sufficient context before the point of the prediction task.

A third possibility is to do hyperparameter tuning of the LSTM. As it stands, we used the hyperparameter values from Li et al. 2017 \cite{li2017code}, but further optimization of these parameters could also improve the performance of the model.

\bibliography{bibliography.bib}

\begin{thebibliography}{}

\bibitem[\protect\astroncite{Allamanis et~al.}{2018}]{allamanis2018survey}
Allamanis, M., Barr, E.~T., Devanbu, P., and Sutton, C. (2018).
\newblock A survey of machine learning for big code and naturalness.
\newblock {\em ACM Computing Surveys (CSUR)}, 51(4):81.

\bibitem[\protect\astroncite{Li et~al.}{2017}]{li2017code}
Li, J., Wang, Y., Lyu, M.~R., and King, I. (2017).
\newblock Code completion with neural attention and pointer networks.
\newblock {\em arXiv preprint arXiv:1711.09573}.

\bibitem[\protect\astroncite{Sennrich et~al.}{2016}]{sennrich-etal-2016-neural}
Sennrich, R., Haddow, B., and Birch, A. (2016).
\newblock Neural machine translation of rare words with subword units.
\newblock In {\em Proceedings of the 54th Annual Meeting of the Association for
  Computational Linguistics (Volume 1: Long Papers)}, pages 1715--1725, Berlin,
  Germany. Association for Computational Linguistics.

\bibitem[\protect\astroncite{Shibata et~al.}{1999}]{shibata1999byte}
Shibata, Y., Kida, T., Fukamachi, S., Takeda, M., Shinohara, A., Shinohara, T.,
  and Arikawa, S. (1999).
\newblock Byte pair encoding: A text compression scheme that accelerates
  pattern matching.
\newblock Technical report, Technical Report DOI-TR-161, Department of
  Informatics, Kyushu University.

\end{thebibliography}
\end{document}